\documentclass[11pt]{article}
\usepackage[utf8]{inputenc}
\usepackage[final]{acl}
\usepackage{multirow, makecell}
\usepackage{amsmath}
\usepackage{algorithm}
\usepackage{algpseudocode}
\usepackage{amssymb}
\usepackage{enumitem}
\usepackage{adjustbox}

\usepackage{times}
\usepackage{latexsym}
\usepackage{graphicx}
\usepackage{xcolor,colortbl}
\usepackage[T1]{fontenc}
\usepackage[utf8]{inputenc}
\usepackage{tcolorbox}
\tcbuselibrary{breakable}
\usepackage{mwe}
\usepackage{tabu}
\usepackage{booktabs}
\usepackage{caption}

\usepackage{microtype}
\usepackage{xurl}
\usepackage{tikz}
\usepackage{pgfplots}
\pgfplotsset{compat=1.18}
\usetikzlibrary{positioning, arrows.meta, fit, calc}

\title{NCL-BU at SemEval-2026 Task 3: Fine-tuning XLM-RoBERTa for Multilingual Dimensional Sentiment Regression}

\author{
  Tong Wu\textsuperscript{1}, Nicolay Rusnachenko\textsuperscript{2}, and Huizhi Liang\textsuperscript{3} \\
  \textsuperscript{1}Independent Researcher\\
  \textsuperscript{2}Centre for Applied Creative Technologies (CFACT+), Bournemouth University, UK \\
  \textsuperscript{3}School of Computing, Newcastle University, Newcastle upon Tyne, UK \\
  \texttt{tongwuwhitney@gmail.com}, 
  \texttt{nrusnachenko@bournemouth.ac.uk}, \\
  \texttt{huizhi.liang@newcastle.ac.uk}
} 

\begin{document}
\maketitle

\begin{abstract}
Dimensional Aspect-Based Sentiment Analysis (DimABSA) extends traditional ABSA from categorical polarity labels to continuous valence--arousal (VA) regression.
This paper describes a system developed for Track~A, Subtask~1 (Dimensional Aspect Sentiment Regression), aiming to predict real-valued VA scores in the $[1, 9]$ range for each given aspect in a text. A fine-tuning approach based on XLM-RoBERTa-base is adopted, with dual regression heads with sigmoid-scaled outputs for valence and arousal prediction. Separate models are trained for each language--domain pair (English and Chinese across restaurant, laptop, and finance domains), and training and development sets are merged for final test predictions. In development experiments, the fine-tuning approach is compared against several large language models under a few-shot prompting setting, demonstrating that task-specific fine-tuning outperforms these LLM-based methods across all evaluation datasets.
\end{abstract}

\section{Introduction}

Aspect-Based Sentiment Analysis (ABSA) aims to identify sentiment expressed toward specific aspects in text and has been extensively studied in the NLP community \citep{zhang2023survey}. Traditional ABSA assigns categorical sentiment labels (e.g., positive, negative, neutral) to each aspect, but such coarse-grained representations fail to capture the nuanced spectrum of human emotion. SemEval-2026 Task~3 - Dimensional ABSA (DimABSA) \citep{yu-etal-2026-semeval, lee2026dimabsabuildingmultilingualmultidomain} bridges this gap by extending ABSA from categorical polarity to continuous valence--arousal (VA) regression based on the circumplex model of affect \citep{russell1980circumplex}. 

This paper focuses on Track~A, Subtask~1 (Dimensional Aspect Sentiment Regression, DimASR), which requires predicting a pair of real-valued VA scores in the range [1,~9] for each given aspect in a text. This work addresses English and Chinese across multiple domains (restaurant, laptop, and finance). The approach fine-tunes XLM-RoBERTa-base \citep{conneau2020unsupervised}, a multilingual pretrained language model, with dual regression heads for valence and arousal prediction. The model takes the concatenation of the input text and the target aspect as input and outputs two real-valued scores via sigmoid-scaled linear heads.

LLM-based approaches have shown strong performance in ABSA classification tasks \citep{zhang-etal-2024-sentiment}, raising the question of whether they can similarly excel at continuous VA regression. To investigate this, a comparative study is conducted between task-specific fine-tuning and few-shot prompting with several state-of-the-art large language models (LLMs). Under the evaluated few-shot prompting conditions, fine-tuning XLM-RoBERTa on modest in-domain annotated data outperforms the evaluated LLM-based methods across all language--domain pairs. The system also outperforms both organizer-provided baselines (Kimi-K2 Thinking and QLoRA-fine-tuned Qwen-3 14B) on all five datasets, with relative improvements of 31--63\%. Error analysis on the test predictions further reveals that the model struggles most with strongly negative sentiment, where it tends to predict values closer to the positive training mean, and that valence is harder to predict than arousal across datasets. The code is available\footnote{\url{https://github.com/tongwu17/SemEval-2026-Task3-Track-A}}.

\section{Background}
\paragraph{Task Description.} This work participates in SemEval-2026 Task~3, Track~A, Subtask~1 \citep{lee2026dimabsabuildingmultilingualmultidomain}. Given a sentence and a target aspect, the system must predict a valence score and an arousal score, both in the range $[1, 9]$. For example, given the sentence ``the food was absolutely amazing!'' and the aspect ``food'', the expected output is $V\!=\!8.50$, $A\!=\!8.25$, indicating positive and excited sentiment. The task provides training data in two languages (English and Chinese) and three domains (restaurant, laptop, and finance).

\paragraph{Aspect-Based Sentiment Analysis.} ABSA has been the focus of a series of SemEval shared tasks \citep{pontiki-etal-2014-semeval, pontiki2015semeval, pontiki-etal-2016-semeval}, progressing from aspect sentiment classification to structured extraction tasks such as Aspect Sentiment Triplet Extraction (ASTE) \citep{peng2020knowing} and Aspect Sentiment Quad Prediction (ASQP) \citep{cai2021aspect, zhang2021aspect}. These tasks, however, represent sentiment as discrete categorical labels.

\paragraph{Dimensional Sentiment Analysis.} In prior work, sentence-level dimensional sentiment resources such as EmoBank \citep{buechel2017emobank} and Chinese EmoBank \citep{lee2022chinese} have enabled valence--arousal prediction. At the aspect level, the SIGHAN~2024 shared task \citep{lee2024sighan} introduced dimensional ABSA for Chinese, and SemEval-2026 Task~3 \citep{lee2026dimabsabuildingmultilingualmultidomain} further extends it to multilingual and multidomain settings.

\paragraph{Pretrained Language Models for Regression.} Transformer-based pretrained models such as BERT \citep{devlin2019bert} produce a [CLS] token representation as a sentence-level encoding that can be augmented with task-specific layers for regression. XLM-RoBERTa \citep{conneau2020unsupervised} extends this paradigm to multilingual settings through large-scale cross-lingual pretraining on 100 languages, outperforming multilingual BERT on cross-lingual benchmarks. This capability makes it well-suited for multilingual dimensional sentiment regression across languages and domains.

\section{Methodology}
\label{sec:methodology}

\subsection{Task Definition}

Given a text $T$ and a set of aspects $\{a_1, a_2, \ldots, a_k\}$, the task is to predict a valence--arousal pair $(V_i, A_i)$ for each aspect $a_i$, where $V_i, A_i \in [1, 9]$.

\subsection{Model Architecture}
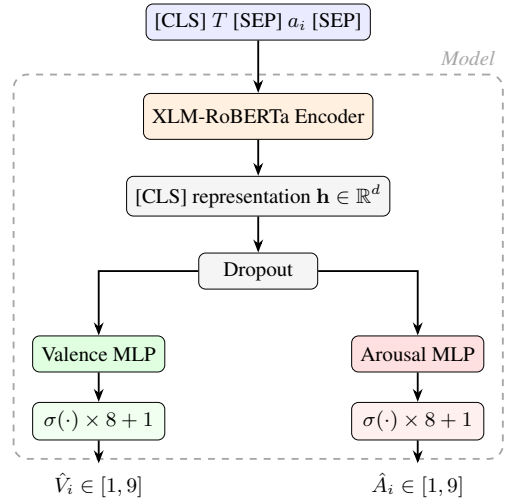
\begin{figure}[t]
\centering
\resizebox{0.85\columnwidth}{!}{%
\begin{tikzpicture}[
    node distance=0.55cm,
    box/.style={rectangle, draw, rounded corners=3pt, minimum width=2.6cm, minimum height=0.55cm, align=center, font=\small},
    headbox/.style={rectangle, draw, rounded corners=3pt, minimum width=2.0cm, minimum height=0.55cm, align=center, font=\small},
    arr/.style={-{Stealth[length=2mm]}, thick},
]
\node[box, fill=blue!8] (input) {$\text{[CLS]}\ T\ \text{[SEP]}\ a_i\ \text{[SEP]}$};
\node[box, fill=orange!12, below=0.8cm of input, minimum width=3.2cm, minimum height=0.7cm] (encoder) {XLM-RoBERTa Encoder};
\node[box, fill=gray!8, below=of encoder] (cls) {[CLS] representation $\mathbf{h} \in \mathbb{R}^d$};
\node[box, fill=gray!8, below=of cls, minimum width=1.8cm] (dropout) {Dropout};
\node[headbox, fill=green!12, below left=0.7cm and 0.5cm of dropout] (v_head) {Valence MLP};
\node[headbox, fill=red!12, below right=0.7cm and 0.5cm of dropout] (a_head) {Arousal MLP};
\node[headbox, fill=green!6, below=0.45cm of v_head] (v_sig) {$\sigma(\cdot) \times 8 + 1$};
\node[headbox, fill=red!6, below=0.45cm of a_head] (a_sig) {$\sigma(\cdot) \times 8 + 1$};
\node[draw, dashed, rounded corners=6pt, gray!70, thick,
      fit=(encoder)(cls)(dropout)(v_head)(a_head)(v_sig)(a_sig),
      inner sep=8pt, label={[font=\small\itshape, gray!70, anchor=south east]north east:Model}] (modelbox) {};
\node[font=\small\bfseries, below=0.35cm of v_sig] (v_out) {$\hat{V}_i \in [1, 9]$};
\node[font=\small\bfseries, below=0.35cm of a_sig] (a_out) {$\hat{A}_i \in [1, 9]$};
\draw[arr] (input) -- (encoder);
\draw[arr] (encoder) -- (cls);
\draw[arr] (cls) -- (dropout);
\draw[arr] (dropout) -| (v_head);
\draw[arr] (dropout) -| (a_head);
\draw[arr] (v_head) -- (v_sig);
\draw[arr] (a_head) -- (a_sig);
\draw[arr] (v_sig) -- (v_out);
\draw[arr] (a_sig) -- (a_out);
\end{tikzpicture}
}%
\caption{Model architecture.}
\label{fig:architecture}
\end{figure}

XLM-RoBERTa-base \citep{conneau2020unsupervised} is used as the backbone encoder. For each (text, aspect) pair, the input is constructed as:
\begin{equation*}
    \text{[CLS]}\ T\ \text{[SEP]}\ a_i\ \text{[SEP]}
\end{equation*}
where [SEP] denotes the separator token of the pretrained tokenizer. The [CLS] token representation $\mathbf{h} \in \mathbb{R}^d$, from the encoder is passed through a dropout layer and then fed into two independent regression heads:
\begin{align*}
    \hat{V}_i &= \sigma(\text{MLP}_V(\mathbf{h})) \times 8 + 1 \\
    \hat{A}_i &= \sigma(\text{MLP}_A(\mathbf{h})) \times 8 + 1
\end{align*}

where $\sigma$ is the sigmoid function and each MLP is a two-layer feedforward network ($d \!\to\! d/2 \!\to\! 1$) with tanh activation and dropout between layers. Since $\sigma$ maps to $[0, 1]$, the affine transformation $\sigma(\cdot) \times 8 + 1$ constrains predictions to $[1, 9]$. Figure~\ref{fig:architecture} illustrates the architecture.

\section{Experimental Setup}
\subsection{Dataset and Training Strategy}

For each language--domain pair, the training data comes from the DimABSA dataset \citep{lee2026dimabsabuildingmultilingualmultidomain}, constructed by annotating VA values on existing ABSA resources. For example, the English restaurant and laptop data are sourced from the ACOS dataset \citep{cai2021aspect}. Since Subtask~1 only requires VA regression for given aspects, (text, aspect, VA) triples are extracted from the \textit{Quadruplet} or \textit{Aspect\_VA} annotations. When a sentence contains multiple aspects, each (text, aspect) pair is treated as an independent sample with its own VA target. Multi-token aspects are kept as raw text spans and tokenized jointly with the sentence by the XLM-RoBERTa tokenizer; no manual subword aggregation is applied to aspect tokens. Training instances are shuffled at the instance level each epoch to reduce any batching bias from repeated sentence contexts shared across multiple aspects. Since each (text, aspect) pair produces a distinct input sequence of the form [CLS]~$T$~[SEP]~$a_i$~[SEP], the encoder representations are not identical across aspects of the same sentence; instance-level shuffling further ensures that aspects from the same sentence are unlikely to co-occur within the same mini-batch.

Table~\ref{tab:dataset} summarizes the dataset statistics. For the final test submission, the training and development sets are merged to maximize the available training data, while a 10\% random split is reserved for internal validation during training.

\begin{table}[t]
\centering
\small
\begin{tabular}{@{}llrrr@{}}
\toprule
\textbf{Language} & \textbf{Domain} & \textbf{Train} & \textbf{Dev} & \textbf{Test} \\
\midrule
eng & Restaurant & 2,284 & 200 & 1,000 \\
eng & Laptop & 4,076 & 200 & 1,000 \\
zho & Restaurant & 6,050 & 300 & 1,000 \\
zho & Laptop & 3,490 & 300 & 1,000 \\
zho & Finance & 1,000 & 200 & 842 \\
\bottomrule
\end{tabular}
\caption{Dataset statistics (number of sentences).}
\label{tab:dataset}
\end{table}

\subsection{Training Configuration}

The training loss is the sum of mean squared error (MSE) for valence and arousal:
\begin{equation*}
    \mathcal{L} = \text{MSE}(\hat{V}, V^*) + \text{MSE}(\hat{A}, A^*)
\end{equation*}
where $V^*$ and $A^*$ are the gold-standard values.

Table~\ref{tab:hyperparams} lists the hyperparameters. Separate models are trained for each language--domain pair: English restaurant, English laptop, Chinese restaurant, Chinese laptop, and Chinese finance.

\begin{table}[t]
\centering
\small
\begin{tabular}{@{}lc@{}}
\toprule
\textbf{Hyperparameter} & \textbf{Value} \\
\midrule
Pretrained model & XLM-RoBERTa-base \\
Max sequence length & 256 \\
Batch size & 16 \\
Learning rate & 2e-5 \\
Optimizer & AdamW \\
Warmup ratio & 10\% \\
LR scheduler & Linear decay \\
Dropout & 0.1 \\
Max epochs & 10 \\
Early stopping patience & 3 \\
Gradient clipping & 1.0 \\
Random seed & 42 \\
\bottomrule
\end{tabular}
\caption{Hyperparameters for fine-tuning.}
\label{tab:hyperparams}
\end{table}

The AdamW optimizer \citep{loshchilov2019decoupled} is used with a linear learning rate schedule including 10\% warmup steps. Early stopping with a patience of 3 epochs is applied based on the RMSE$_{VA}$ metric on a held-out validation split (10\% of training data). All results reflect a single training run per language--domain pair with a fixed random seed (42).

\subsection{Development Evaluation Protocol}
To compare methods during development, the original training set is split into 80\% for training and 20\% for evaluation. This setup allows comparison of the fine-tuning approach against LLM-based methods on a held-out portion of annotated data.

\subsection{Evaluation Metric}
The evaluation metric for Subtask~1 is RMSE$_{VA}$, defined as:
{\small
\begin{equation*}
    \text{RMSE}_{VA} = \sqrt{\frac{1}{N}\sum_{i=1}^{N} \left[(V_p^{(i)} - V_g^{(i)})^2 + (A_p^{(i)} - A_g^{(i)})^2\right]}
\end{equation*}
}%
where $N$ is the total number of (text, aspect) instances, and subscripts $p$ and $g$ denote predicted and gold values, respectively. Lower values indicate better performance.

\section{Results}

\subsection{Main Results}
Models trained on the training set alone, without using any development gold labels for training, achieve RMSE$_{VA}$ of 1.1045, 1.0630, 0.7207, 0.7983, and 0.5389 on the development set for eng\_lap, eng\_res, zho\_lap, zho\_res, and zho\_fin, respectively. For the final test predictions, the training and development sets are merged to maximize available data, with an internal 10\% validation split reserved for early stopping.

Table~\ref{tab:submission_results} reports the resulting test scores alongside the organizer-provided baselines. Among the baselines, Kimi-K2 Thinking achieves lower RMSE$_{VA}$ on four of five datasets, while Qwen-3 14B leads only on Chinese Finance (1.4707 vs.\ 1.9652). XLM-RoBERTa outperforms both baselines on all five datasets, with improvements over the per-dataset stronger baseline ranging from 31\% to 63\% across the five language--domain pairs. The largest gains occur on Chinese datasets, where relative improvements reach 50--63\%, notably higher than the 31--34\% gains on English datasets, where all three Chinese datasets achieve RMSE$_{VA}$ below 1.0 while both baselines remain above 1.4.

\begin{table}[t]
\centering
\small
\resizebox{\columnwidth}{!}{%
\begin{tabular}{@{}lccccc@{}}
\toprule
\textbf{Method} & \textbf{eng\_lap} & \textbf{eng\_res} & \textbf{zho\_lap} & \textbf{zho\_res} & \textbf{zho\_fin} \\
\midrule
Kimi-K2 Thinking & 2.1893 & 2.1461 & 1.6440 & 1.8959 & 1.9652 \\
Qwen-3 14B (QLoRA) & 2.8089 & 2.6427 & 1.7706 & 2.0073 & 1.4707 \\
\midrule
\textbf{XLM-RoBERTa} & \textbf{1.4562} & \textbf{1.4861} & \textbf{0.7510} & \textbf{0.9553} & \textbf{0.5391} \\
\bottomrule
\end{tabular}%
}
\caption{Test set RMSE$_{VA}$ compared with baselines (lower is better).}
\label{tab:submission_results}
\end{table}

\subsection{Training Convergence Analysis}

\begin{figure}[t]
\centering
\begin{tikzpicture}
\begin{axis}[
    width=\columnwidth,
    height=6.2cm,
    xlabel={Epoch},
    ylabel={RMSE$_{VA}$},
    xmin=0.5, xmax=10.5,
    ymin=0.3, ymax=2.3,
    xtick={1,2,3,4,5,6,7,8,9,10},
    legend style={at={(0.96,0.96)}, anchor=north east, font=\scriptsize, cells={anchor=west}, row sep=-2pt},
    ymajorgrids=true,
    grid style={dashed, gray!30},
    tick label style={font=\small},
    label style={font=\small},
]
\addplot[color=blue, mark=*, thick, mark size=1.5pt] coordinates {
    (1, 1.2790) (2, 1.1368) (3, 1.1403) (4, 1.1847) (5, 1.1463)
};
\addplot[color=red, mark=square*, thick, mark size=1.5pt] coordinates {
    (1, 1.9628) (2, 1.4452) (3, 1.2767) (4, 1.1734) (5, 1.1450) (6, 1.1846) (7, 1.1757) (8, 1.1262) (9, 1.1229) (10, 1.1353)
};
\addplot[color=green!60!black, mark=triangle*, thick, mark size=1.5pt] coordinates {
    (1, 1.1213) (2, 0.9518) (3, 0.8191) (4, 0.8762) (5, 0.8315) (6, 0.8163) (7, 0.8328) (8, 0.8175) (9, 0.8319)
};
\addplot[color=orange, mark=diamond*, thick, mark size=1.5pt] coordinates {
    (1, 0.8037) (2, 0.7487) (3, 0.7518) (4, 0.7085) (5, 0.6575) (6, 0.6738) (7, 0.6688) (8, 0.6753)
};
\addplot[color=purple, mark=pentagon*, thick, mark size=1.5pt] coordinates {
    (1, 0.6117) (2, 0.6236) (3, 0.6043) (4, 0.4750) (5, 0.4449) (6, 0.4487) (7, 0.4464) (8, 0.4687)
};
\legend{eng\_lap, eng\_res, zho\_lap, zho\_res, zho\_fin}
\end{axis}
\end{tikzpicture}
\caption{Validation RMSE$_{VA}$ across training epochs.}
\label{fig:training_curves}
\end{figure}
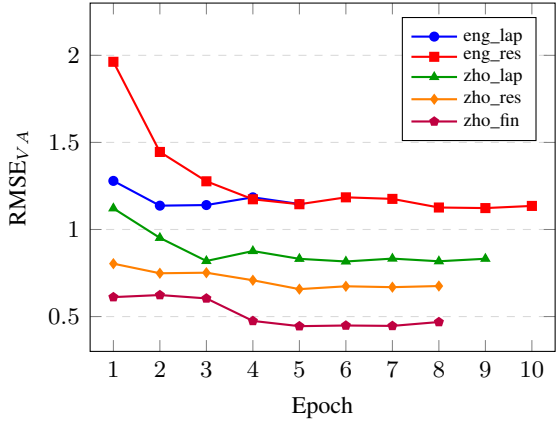

Figure~\ref{fig:training_curves} shows the validation RMSE$_{VA}$ across training epochs for all five datasets when trained on the merged train+dev data. Several observations can be made: (1)~Chinese datasets converge to lower error levels than English ones, consistent with the larger training sizes; (2)~early stopping effectively prevents overfitting, with most models stopping between epochs 5--9; (3)~English Restaurant requires the most epochs (10) to converge, likely due to its smaller training set combined with higher annotation variability.

\subsection{Comparison with LLM-based Methods}

To compare against prompting-based approaches, several LLMs are evaluated under a few-shot setting on the same 20\% held-out evaluation set:

\begin{itemize}[nosep]
    \item \textbf{GPT-5.2}: A proprietary OpenAI model, accessed via API.
    \item \textbf{LLaMA-3-70B-Instruct}: A 70B-parameter open-source model by Meta.
    \item \textbf{LLaMA-4-Maverick-Instruct}: A 400B-parameter mixture-of-experts model by Meta.
\end{itemize}

For the LLM methods, a structured prompt is used with system instructions defining valence and arousal, followed by 6 few-shot examples sampled from the 80\% training portion. A low temperature of 0.1 is used to encourage deterministic outputs. The prompt asks the model to output VA scores in the format V\#A. Predicted values are clipped to the [1,~9] range. The same prompt format and few-shot examples are applied uniformly across all LLMs and datasets. Table~\ref{tab:dev_results} presents the results.

\begin{table}[t]
\centering
\small
\resizebox{\columnwidth}{!}{%
\begin{tabular}{@{}lccccc@{}}
\toprule
\textbf{Method} & \textbf{eng\_lap} & \textbf{eng\_res} & \textbf{zho\_lap} & \textbf{zho\_res} & \textbf{zho\_fin} \\
\midrule
GPT-5.2 & 1.9663 & 1.9203 & 1.5825 & 1.7279 & 1.7311 \\
LLaMA-3-70B & 1.6243 & 1.6748 & 1.9249 & 1.9589 & 2.2312 \\
LLaMA-4-Mav. & 1.6987 & 1.7955 & 1.6480 & 1.7718 & 1.7858 \\
\midrule
\textbf{XLM-RoBERTa} & \textbf{1.1084} & \textbf{1.3402} & \textbf{0.8034} & \textbf{0.6727} & \textbf{0.5118} \\
\bottomrule
\end{tabular}%
}
\caption{RMSE$_{VA}$ comparison on the 20\% held-out evaluation set (lower is better).}
\label{tab:dev_results}
\end{table}

\begin{figure}[t]
\centering
\includegraphics[width=\columnwidth]{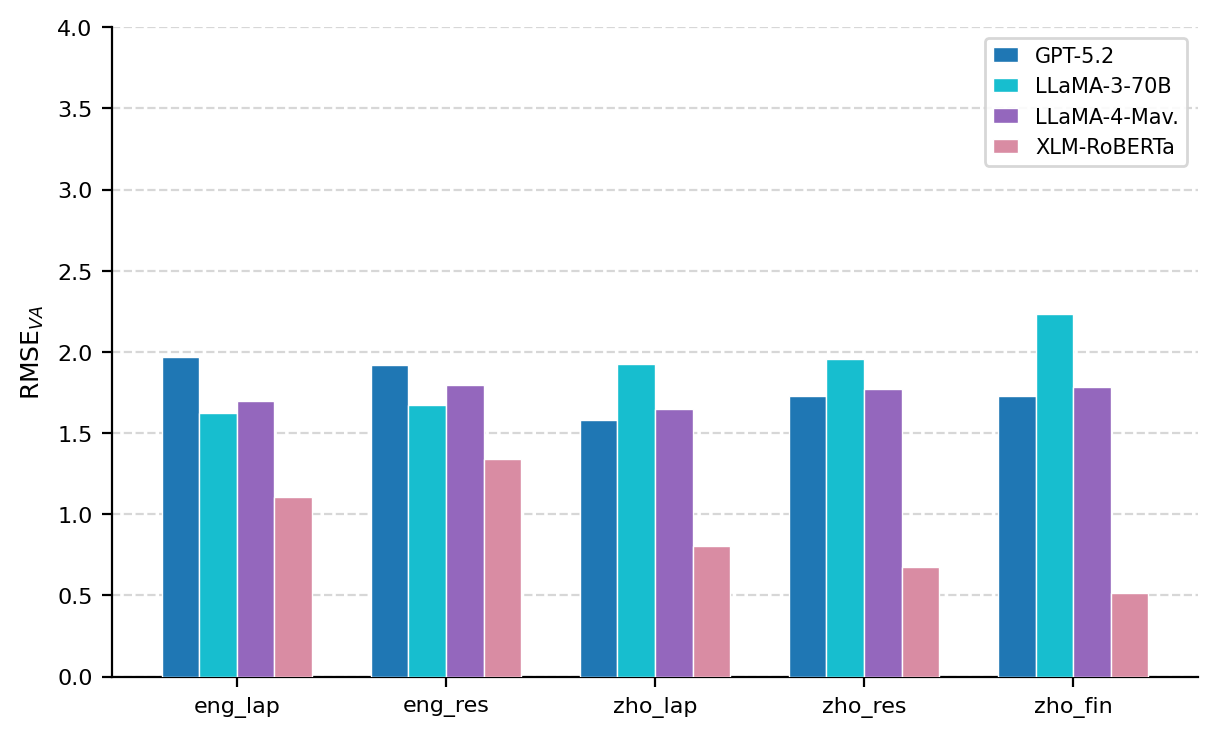}
\caption{Visual comparison of RMSE$_{VA}$ across methods on the 20\% held-out evaluation set.}
\label{fig:llm_comparison}
\end{figure}

\paragraph{Fine-tuning vs.\ LLMs.} Under the few-shot prompting setup, the fine-tuned XLM-RoBERTa models outperform the evaluated LLM-based methods across all datasets. Figure~\ref{fig:llm_comparison} makes this gap visually explicit: the XLM-RoBERTa bars are consistently the lowest in every language--domain pair. The average improvement over the best LLM per dataset is 0.78 RMSE$_{VA}$ points (a relative reduction of approximately 46\%). The performance gap is particularly striking for Chinese datasets, where the best LLM achieves RMSE$_{VA}$ of 1.5825 on Chinese Laptop while the fine-tuned model achieves 0.8034, and similarly 1.7279 vs.\ 0.6727 on Chinese Restaurant.

\paragraph{LLM Comparison.} Among the LLMs, no single model dominates across all datasets. Within the prompting-based group, LLaMA-3-70B achieves the lowest RMSE$_{VA}$ on both English datasets, whereas GPT-5.2 performs best on all three Chinese datasets, suggesting language-specific calibration differences. The gap between the best- and worst-performing LLM can be substantial (e.g., 1.7311 to 2.2312 on Chinese Finance), suggesting that VA regression performance varies considerably across models and languages.

\paragraph{Analysis of LLM Limitations.} The substantial gap between LLM-based approaches and fine-tuning can be attributed to several factors: (1)~VA regression requires predicting precise numerical values on a continuous scale, which is inherently challenging for LLMs that generate text token by token; (2)~the few-shot setting provides very limited calibration signal (only 6 examples) for the model to learn the dataset-specific distribution; (3)~fine-tuning directly optimizes MSE loss, providing a clear gradient signal for regression, while LLMs must infer numerical patterns from text.

\subsection{Error Analysis}

\begin{figure}[t]
\centering
\includegraphics[width=0.88\columnwidth]{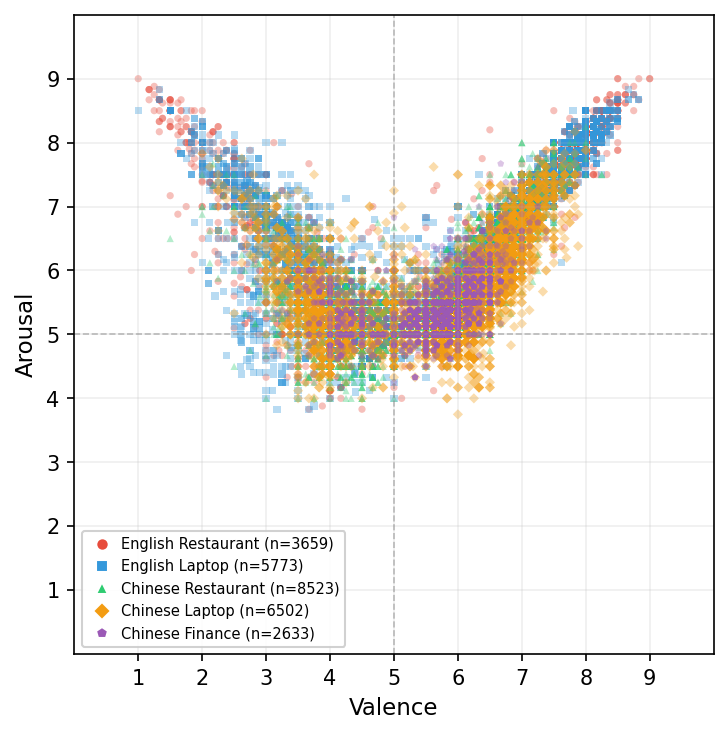}
\caption{Training data distribution in VA space. English datasets cluster in the high-valence, high-arousal region, while Chinese datasets are more centered.}
\label{fig:va_distribution}
\end{figure}

\begin{figure*}[t]
\centering
\includegraphics[width=0.92\textwidth]{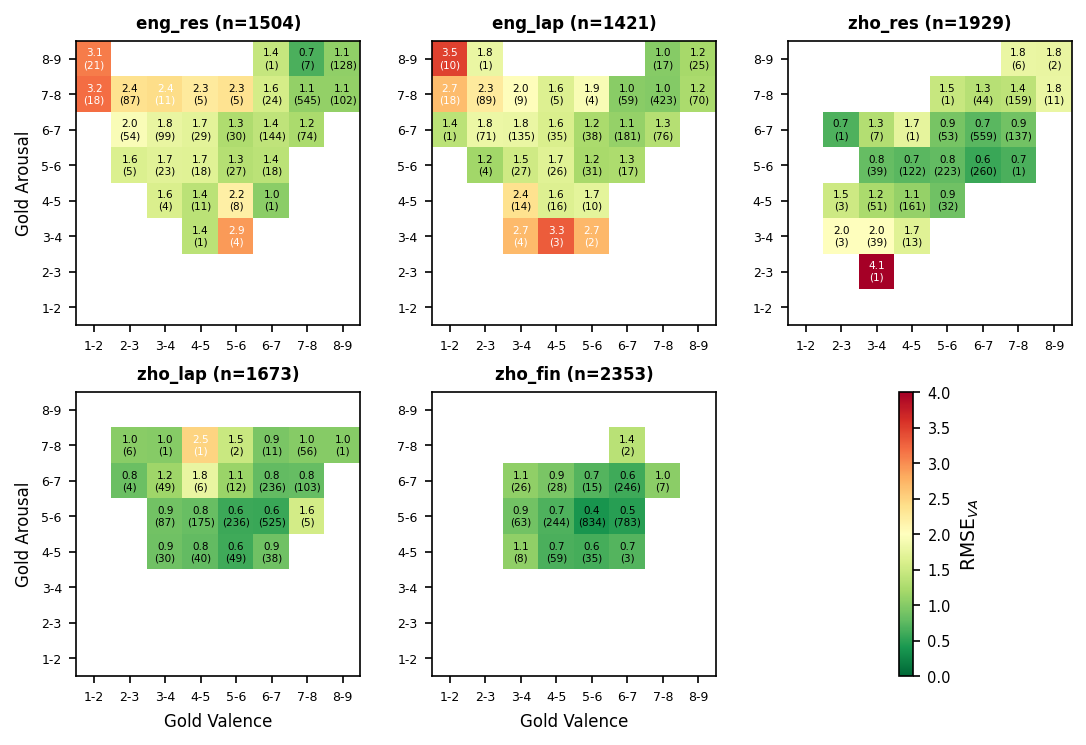}
\caption{Test-set RMSE\textsubscript{VA} heatmap across the VA space. Each cell shows the RMSE and sample count~(n) for gold samples in that valence--arousal bin. Red indicates higher error; green indicates lower error.}
\label{fig:error_heatmap}
\end{figure*}

\paragraph{Cross-lingual Observations.} The fine-tuned model performs better on Chinese datasets than on English ones (average test RMSE$_{VA}$ of 0.7485 vs.\ 1.4712 for Chinese and English, respectively). This may be attributed to the larger training sets available for Chinese domains (e.g., 6,050 sentences for Chinese Restaurant); annotation consistency differences between languages are plausible but have not been directly verified in this study. Chinese Finance achieves the lowest error (0.5391). The zho\_fin annotations are more uniformly distributed across the VA space than the other four datasets, producing the smallest per-dimension errors on both valence (0.44) and arousal (0.31) in Table~\ref{tab:rmse_breakdown}.

\paragraph{Distributional Bias.} Figure~\ref{fig:va_distribution} shows the training data distribution in the VA space across all five datasets. English datasets are heavily skewed toward positive valence and high arousal (mean $V\!\approx\!6.1$, $A\!\approx\!6.8$), while Chinese datasets are more centered (mean $V\!\approx\!5.7$, $A\!\approx\!5.5$). This distributional imbalance directly explains several test prediction patterns, as the model tends to predict values closer to the positive training mean, failing to capture extreme negativity. For example, for the sentence ``If the line is longer I would highly recommend skipping this place and heading down the street to Fido -- insanely good coffee and great breakfast food,'' the model predicts nearly identical valence for all three aspects (\textit{place}: $V\!=\!7.38$; \textit{coffee}: $V\!=\!7.43$; \textit{breakfast food}: $V\!=\!7.71$), while gold labels differ substantially ($V\!=\!2.88$, $8.38$, and $7.88$, respectively). The overall positive tone of the sentence (``highly recommend'', ``insanely good'') dominates the encoding, causing the model to miss the negative sentiment toward \textit{place}.
As this example illustrates, although each aspect is individually concatenated to the input, the prediction relies on the [CLS] representation, which is a global mixture of all input tokens via self-attention. Without a dedicated mechanism to focus on tokens relevant to the target aspect, the positive tone of surrounding context can dominate this global encoding regardless of which aspect is being evaluated, causing the model to assign near-identical predictions to aspects with substantially different gold values. This failure is a direct consequence of the [CLS]-based architecture: replacing global pooling with aspect-span attention or explicit aspect markers in the input would be a natural architectural response to such over-smoothing errors.

Figure~\ref{fig:error_heatmap} provides a spatially resolved view of these patterns. For English datasets, the highest errors concentrate in the low-valence region ($V\!<\!3$), with RMSE$_{VA}$ of 2.43 for restaurant and 2.20 for laptop, compared to only 1.09 and 1.08 in the high-valence region $[7, 9]$, confirming the distributional bias. Chinese datasets exhibit more uniform error surfaces, with considerably less extreme variation across the entire VA space, consistent with their more balanced training distributions. Notably, the heatmap cells with the fewest training samples (low-valence, high-arousal) consistently show the largest errors, reinforcing that data scarcity in underrepresented VA regions is the primary driver of prediction errors.

Furthermore, valence is generally harder to predict than arousal across language--domain pairs. Table~\ref{tab:rmse_breakdown} shows the per-dimension RMSE: valence RMSE exceeds arousal RMSE in four out of five datasets, with the largest gap on English Laptop (1.25 vs.\ 0.75); the sole exception is Chinese Restaurant, where arousal error is higher (0.74 vs.\ 0.61). This suggests that arousal, being more closely tied to surface-level cues such as exclamation marks and intensifiers, is easier for the model to capture, while valence requires deeper semantic understanding of sentiment polarity.

\begin{table}[t]
\centering
\small
\begin{tabular}{@{}llcc@{}}
\toprule
\textbf{Language} & \textbf{Domain} & \textbf{RMSE\textsubscript{V}} & \textbf{RMSE\textsubscript{A}} \\
\midrule
eng & Laptop & 1.25 & 0.75 \\
eng & Restaurant & 1.11 & 0.98 \\
zho & Laptop & 0.57 & 0.48 \\
zho & Restaurant & 0.61 & 0.74 \\
zho & Finance & 0.44 & 0.31 \\
\bottomrule
\end{tabular}
\caption{Test-set RMSE for valence (RMSE\textsubscript{V}) and arousal (RMSE\textsubscript{A}).}
\label{tab:rmse_breakdown}
\end{table}

\section{Conclusion}

This paper presented a fine-tuning approach based on XLM-RoBERTa-base with dual regression heads for dimensional aspect sentiment regression. Experiments on five language--domain pairs demonstrate that task-specific fine-tuning outperforms few-shot prompting with several LLMs under the evaluated conditions. Valence is generally harder to predict than arousal. The error analysis identifies key error patterns: [CLS]-based over-smoothing for co-occurring aspects, and distributional bias toward positive training values. Incorporating these insights into an improved system was precluded by shared-task time constraints and the need to preserve comparability with submitted results; aspect-aware pooling and data augmentation in data-sparse VA regions remain natural directions for future work. Future work includes using larger pretrained models, multi-task learning across subtasks, and incorporating affective lexicons for low-resource domains.

\section*{Limitations}

The model uses only the [CLS] representation without aspect-level attention. The LLM comparison uses a single prompting setup with 6 few-shot examples, which may not fully represent LLM capabilities. The evaluation covers only two languages, leaving generalizability untested. Each language--domain model is also trained independently without parameter sharing. Joint or multi-task training across domains and languages may improve data efficiency, particularly for smaller datasets such as Chinese Finance. No ablation studies were conducted comparing, e.g., a single regression head with a 2-dimensional output against two independent scalar heads, or aspect-aware pooling against plain [CLS] encoding; such comparisons are left for future work.

\section*{Acknowledgments}
Thanks to the organizers for providing the dataset and evaluation infrastructure.

\bibliography{task3-ref} 

\appendix
\vspace{-0.5\baselineskip}
\section{Prompt}
\label{sec:appendix_prompt}

The following prompt is used for all LLM-based comparison methods. It consists of a system prompt and a set of 6 few-shot examples.

\subsection{System Prompt}

\begin{tcolorbox}[colback=blue!3, colframe=blue!40, title=System Prompt, fonttitle=\bfseries, breakable]
You are an expert in sentiment analysis. Your task is to predict Valence and Arousal scores for aspects in sentences.

\medskip
\textbf{Definitions:}
\begin{itemize}[nosep, leftmargin=*]
    \item \textbf{Valence}: emotional positivity/negativity (1.0 = very negative, 5.0 = neutral, 9.0 = very positive)
    \item \textbf{Arousal}: emotional intensity/excitement (1.0 = very calm/sluggish, 5.0 = moderate, 9.0 = very excited)
\end{itemize}

\medskip
\textbf{Output format:} valence\#arousal (e.g., 7.50\#6.80)
\end{tcolorbox}

\subsection{Few-shot Examples}

\begin{tcolorbox}[colback=gray!5, colframe=gray!50, title=Few-shot Examples, fonttitle=\bfseries, breakable]
\begin{enumerate}[nosep, leftmargin=*]
    \item \textbf{Text:} ``the food was absolutely amazing!!'' \\
          \textbf{Aspect:} ``food'' \\
          \textbf{Answer:} 8.50\#8.25

    \item \textbf{Text:} ``but the staff was so horrible to us.'' \\
          \textbf{Aspect:} ``staff'' \\
          \textbf{Answer:} 1.33\#8.67

    \item \textbf{Text:} ``food was just average... if they lowered the prices just a bit, it would be a bigger draw.'' \\
          \textbf{Aspect:} ``food'' \\
          \textbf{Answer:} 5.00\#5.00

    \item \textbf{Text:} ``i love this macbook.'' \\
          \textbf{Aspect:} ``macbook'' \\
          \textbf{Answer:} 7.10\#6.90

    \item \textbf{Text:} ``horrible product.'' \\
          \textbf{Aspect:} ``product'' \\
          \textbf{Answer:} 2.60\#5.70

    \item \textbf{Text:} ``it has and does everything it should.'' \\
          \textbf{Aspect:} ``NULL'' \\
          \textbf{Answer:} 5.67\#5.50
\end{enumerate}
\end{tcolorbox}

\section{Prediction Examples}
\label{sec:appendix_examples}

Table~\ref{tab:examples} shows selected test predictions. Near-exact cases match gold values closely; failure cases involve sarcasm or implicit negativity where the model predicts positive values.

\vspace{0.5\baselineskip}
\begin{center}
\small
\setlength{\tabcolsep}{4pt}
\begin{tabular}{@{}p{4.5cm}ccc@{}}
\toprule
\textbf{Text (Aspect)} & & \textbf{Pred} & \textbf{Gold} \\
\midrule
\multicolumn{4}{@{}l}{\textit{Near-exact predictions}} \\[2pt]
``I enjoy real flavor, real fruits'' (\textit{flavor}) & V & 7.75 & 7.75 \\
 & A & 7.75 & 7.75 \\[3pt]
``The keyboard is full size and the spacing \ldots\ comfortable'' (\textit{keyboard}) & V & 6.99 & 7.00 \\
 & A & 7.02 & 7.00 \\[3pt]
``We shared the big easy breakfast and it was okay'' (\textit{big easy breakfast}) & V & 6.49 & 6.50 \\
 & A & 6.31 & 6.33 \\[3pt]
``The battery still lasts about 3.5 hours \ldots'' (\textit{battery}) & V & 5.99 & 6.00 \\
 & A & 5.99 & 6.00 \\
\midrule
\multicolumn{4}{@{}l}{\textit{Failure cases}} \\[2pt]
``I ordered a burger medium and got a charred, tasteless hockey puck'' (\textit{burger}) & V & 4.03 & 1.38 \\
 & A & 5.22 & 8.50 \\[3pt]
``My wife also had the pleasure of adding spoiled creamer'' (\textit{creamer}) & V & 6.80 & 1.67 \\
 & A & 6.75 & 8.17 \\[3pt]
``I'm being generous by giving this restaurant 2 stars'' (\textit{restaurant}) & V & 7.08 & 2.17 \\
 & A & 7.08 & 7.67 \\[3pt]
``Let me note that this waitress is giving the customers at the 3 tables surrounding us WAY better customer service'' (\textit{waitress}) & V & 5.84 & 1.83 \\
 & A & 5.82 & 8.00 \\
\bottomrule
\end{tabular}
\captionof{table}{Selected test predictions.}
\label{tab:examples}
\end{center}

\section{Error Distribution}
\label{sec:appendix_error_dist}

Table~\ref{tab:error_dist} shows the distribution of per-instance RMSE$_{VA}$ errors on the test set, where each error is the Euclidean distance between predicted and gold VA points on the $[1, 9]$ scale. Over 94\% of Chinese Finance predictions fall below 1.0 (accurate), while English datasets have 16--17\% with error above 2.0 (large errors); the median error remains below 1.0 across all three Chinese datasets.

\vspace{0.5\baselineskip}
\noindent\begin{minipage}{\columnwidth}
\centering
\small
\begin{tabular}{@{}llccc@{}}
\toprule
\textbf{Lang} & \textbf{Domain} & \textbf{Median} & \textbf{\%$<$1.0} & \textbf{\%$>$2.0} \\
\midrule
eng & Laptop & 0.87 & 55.6 & 16.9 \\
eng & Restaurant & 0.78 & 60.1 & 16.2 \\
zho & Laptop & 0.44 & 86.6 & 2.2 \\
zho & Restaurant & 0.66 & 72.2 & 3.4 \\
zho & Finance & 0.38 & 94.8 & 0.1 \\
\bottomrule
\end{tabular}
\phantomsection
\captionof{table}{Per-instance RMSE$_{VA}$ error distribution on the test set.}\label{tab:error_dist}
\end{minipage}

\end{document}